\documentclass[conference]{IEEEtran}
\IEEEoverridecommandlockouts
\usepackage{amsmath,amssymb,amsfonts}
\usepackage{graphicx}
\usepackage{textcomp}
\usepackage{xcolor}
\usepackage{booktabs}
\usepackage{multirow}
\usepackage{amsmath,amssymb}
\usepackage{algorithm}
\usepackage{algpseudocode}
\usepackage{subfigure}
\usepackage{ulem}
\def\BibTeX{{\rm B\kern-.05em{\sc i\kern-.025em b}\kern-.08em
    T\kern-.1667em\lower.7ex\hbox{E}\kern-.125emX}}
\begin{document}

\title{Fed-Listing: Federated Label Distribution Inference in Graph Neural Networks
}

\author{\IEEEauthorblockN{
		Suprim Nakarmi\IEEEauthorrefmark{1},
		Junggab Son\IEEEauthorrefmark{1},
		Yue Zhao\IEEEauthorrefmark{2} and
		Zuobin Xiong\IEEEauthorrefmark{1}}
		\IEEEauthorblockA{
		\IEEEauthorrefmark{1}Department of Computer Science, University of Nevada Las Vegas, Las Vegas, USA\\
		\IEEEauthorrefmark{2}Department of Computer Science, University of Southern California, Los Angeles, USA\\
		\IEEEauthorrefmark{1} {\it nakars2@unlv.nevada.edu, \{junggab.son, zuobin.xiong\}@unlv.edu};
		\IEEEauthorrefmark{2} {\it yue.z@usc.edu}}
		}

\maketitle

\begin{abstract}
% gnn background
% Graph Neural Networks (GNNs) have been intensively studied for their expressive representation and learning performance on graph-structured data, enabling effective modeling of complex relational dependencies among nodes and edges in various domains such as social networks, molecular chemistry, and recommendation systems.
% % current caveats 
% However, the standalone GNNs can unleash threat surfaces and privacy implications, as some sensitive graph-structured data is collected and processed in a centralized setting. 
% To solve this issue and achieve data autonomy, 
Federated Graph Neural Networks (FedGNNs) facilitate collaborative learning across multiple clients with graph-structured data while preserving user privacy. 
However, emerging research indicates that within this setting, shared model updates, particularly gradients, can unintentionally leak sensitive information of local users. 
% motivation
Numerous privacy inference attacks have been explored in traditional federated learning and extended to graph settings, but the problem of label distribution inference in FedGNNs remains largely underexplored.
In this work, we introduce Fed-Listing (\underline{Fed}erated \underline{L}abel D\underline{ist}ribution \underline{In}ference in \underline{G}NNs), a novel gradient-based attack designed to infer the private label statistics of target clients in FedGNNs without access to raw data or node features. 
Fed-Listing only leverages the final-layer gradients exchanged during training to uncover statistical patterns that reveal class proportions in a stealthy manner. 
% An auxiliary shadow dataset is used to generate diverse label partitioning strategies, simulating various client distributions, on which the attack model is obtained. 
Extensive experiments on four benchmark datasets and three GNN architectures show that Fed-Listing significantly outperforms existing baselines, including random guessing and Decaf, even under challenging non-i.i.d. scenarios. 
Moreover, existing defense mechanisms can barely reduce the attack performance of Fed-Listing, unless the model's utility is severely degraded.
% {\color{red}can existing defense mitigate our attack?}
% Our findings expose a critical privacy vulnerability in FedGNNs and highlight the need for more effective defenses against gradient-based information leakage. 
The \emph{code implementation} and \emph{Supplementary materials} are available here: https://github.com/suprimnakarmi/Fed-Listing.
\end{abstract}


\section{Introduction}
% GNN background
Graph Neural Networks (GNNs) have emerged as powerful tools for learning from graph-structured data, where both nodes and edges provide essential information for capturing relational patterns for various tasks. 
By leveraging message passing and neighborhood aggregation mechanisms, GNNs can capture complex dependencies among nodes in diverse applications such as social networks~\cite{ guo2020deep, sharma2024survey}, molecular graphs~\cite{wang2023graph, zhang2021graph}, traffic networks{~\cite{jiang2023graph, sharma2023graph}}, and recommendation systems{~\cite{gao2022graph, wu2022graph}}.
However, traditional GNNs training assumes centralized access to the entire graph, which poses significant privacy and scalability challenges when data is distributed across multiple organizations or edge devices.

% GNN to FedGNN
Federated Graph Neural Networks (FedGNNs) extend the principles of Federated Learning (FL) to graph-based domains, enabling decentralized and privacy-preserving model training without directly sharing raw data~\cite{liu2024federated}. 
In FedGNNs, individual participants (clients) compute local model updates on their private graph data and only share the model weights or gradients with a central server, which aggregates them to update the global model. 
This setup is particularly important in privacy-sensitive domains, where data cannot be centralized due to legal, ethical, or security constraints{~\cite{shiranthika2023decentralized, zheng2023decentralized, zhang2024systematic}}. 
Real-world use cases include personalized recommendation systems across e-commerce platforms \cite{liu2022federated}, collaborative medical image analysis across hospitals \cite{ahmed2025fedgraphmri,balik2022investigating}, and cross-institutional drug discovery research \cite{manu2024graphganfed}.

% Challenges of FedGNN
Despite its promise, FedGNNs inherit common problems of FL mechanisms and introduce unique challenges due to the interplay between different graph topologies. 
First, recent studies have shown that even when only the model parameters are shared in FL, they may still leak sensitive information. 
These shared parameters often retain latent footprints of the underlying private data and are susceptible to manipulation or inversion attacks that can partially or fully reconstruct the original dataset~\cite{geiping2020inverting, yang2023gradient}. 
Moreover, FedGNNs, in particular, inherit and amplify these vulnerabilities due to the rich relational structure encoded in graphs.
Different attacks have been identified in the FedGNNs literature, including membership inference attacks, where adversaries determine whether a particular data point was part of the training set~\cite{olatunji2021membership, bai2024membership}; 
property inference attacks, which aim to infer global attributes or properties of the client’s private dataset~\cite{liu2025piafgnn}; 
graph reconstruction attacks, where attackers attempt to reconstruct nodes, edges, or entire graphs based on shared information~\cite{drencheva2025grain}; 
and adversarial attacks that manipulate model updates to degrade global performance or mislead predictions~\cite{sun2022adversarial}.
These attack vectors are well-documented and reflect the importance and growing demand for FedGNN studies.
Therefore, understanding the security and privacy implications of federated training on graph data is essential for ensuring robust deployment in sensitive real-world systems. 
In this work, we investigate an emerging privacy attack in FedGNNs -- the Label Distribution Inference (LDI) attack, which has been discussed in a centralized scenario, yet remains underexplored in the FedGNNs settings. 

% Although federated learning (FL) is designed to prevent raw data sharing, gradients and model updates may still encode rich distributional information about local datasets. 

Leakage of label distribution can expose sensitive information such as demographic traits, user shopping behaviors, or medical conditions in healthcare graphs, leading to privacy violations, competitive disadvantages, and potential downstream preference attacks.
In graph data, this risk is amplified due to homophily, where nodes of similar labels tend to connect, causing structural and feature correlations to implicitly reveal community composition and client-specific subgraph semantics. 
Building upon these insights, this paper introduces a practical threat model where the federated server aims to infer the underlying label distribution of a target client, i.e., deducing the statistical distribution of class labels in the target client’s private dataset.
% To the best of our knowledge, despite the pressing concerns on privacy leakage and downstream model manipulation, the LDI in FedGNNs models is still an open question. 
% LDI aims to uncover the proportion of each class within a client’s dataset rather than just the presence or absence of specific labels. 
This exposes highly sensitive information, for instance, estimating the number of patients with tumor-positive versus normal scans in a hospital, or identifying dominant and minority product lines in a retailer’s sales data.
Such information can give adversaries an indirect understanding of client data, with a significant impact on privacy and competitive confidentiality for further preference attack and even user profiling. 

To launch LDI in FedGNNs, we propose Fed-Listing, an attack framework tailored for horizontal FedGNN settings.
Fed-Listing leverages an auxiliary dataset for multiple shadow federated training instances under diverse client distribution strategies. 
From these shadow trainings, we construct an attack dataset by recording gradients of the shadow client parameters, which are then used to train a neural network–based attack model. 
To validate our attack, we evaluate Fed-Listing on four graph datasets and three widely adopted defense strategies. 
Our results show that while these defenses mitigate leakage to varying degrees, Fed-Listing consistently outperforms baselines, such as random guessing and Decaf \cite{dai2024decaf}, particularly in challenging distribution scenarios, including single-class or one-class-dominant distributions.

Our contributions can be summarized as follows:
\begin{enumerate}
    \item We propose Fed-Listing, the first passive and stealth LDI attack specifically for the horizontal FedGNNs setting.
    \item The designed shadow-training pipeline in Fed-Listing using an auxiliary dataset explores heterogeneous data distributions and creates diverse attack scenarios.
    \item Extensive evaluations are conducted on four benchmark graph datasets and three popular GNN architectures, demonstrating the superior performance of Fed-Listing.
    \item We empirically analyze the resilience of Fed-Listing against three widely adopted defense mechanisms, providing insights on the trade-off between privacy and model utility.
\end{enumerate}

% The remainder of the paper is organized as follows. 
% We review the related literature and preliminary knowledge in Section~\ref{sec:related} and Section~\ref{sec:pre}. 
% After presenting the model framework in Section~\ref{sec:fed-listing}, we design comprehensive experiments and exhibit results in Section~\ref{sec:exp}. 
% To fully understand the method, an ablation study is conducted in Section~\ref{sec:ablation}. 
% Finally, this paper is concluded in Section~\ref{sec:conclusion}.

\section{Related work}
\label{sec:related}
Existing literature related to this work can be categorized into two main branches: label inference attacks and LDI attacks. 

\subsection{Label Inference Attacks}
Several studies focus on identifying the presence of specific class labels in a client’s dataset within the FL system.
For instance, Wainakh et al. proposed Label Leakage by Gradient (LLG), which exploits gradients from the last layer in the federated setting to detect whether a particular label exists in a client’s local training data~\cite{wainakh2021user}. 
Similarly, {Meng et al.} leverages the message passing mechanism of GNN by adding a single infiltrator (or fake) node to the victim node and use posterior outputs to infer the presence of label~\cite{meng2023devil}. 
{Arazzi et al.} performed a label inference attack in Vertical Federated Learning (VFL) by initializing synthetic labels and iteratively updating them, along with a server model approximation, to minimize the difference between adversarial gradients they generated and the real gradients returned by the server \cite{arazzi2023label}.
% Their method filters out labels with negative gradients and calibrates the remaining gradients by removing a per-label offset, iteratively selecting the most negative gradient to infer labels within a known batch size.
% 
% \subsection{Preference Profiling Attacks}
Apart from exact label inference, a variant attack targets the preference of a client, referred to as the label profiling attack. 
{Zhou et al.} were the first to introduce the Profiling Preference Attack (PPA), exploiting per-class gradient sensitivity to infer a user’s dominant and minority classes \cite{zhou2022ppa}, in which they trained a meta-classifier that predicts client preferences based on gradient sensitivity across classes. 
Extending this line, {Liu et al.} proposed Surrogate Generation Preference Profiling (SGPP), a graph-based PPA tailored for VFL, which infers class preferences using only a trained extractor combined with a domain adaptation strategy~\cite{liu2025preference}.

In summary, existing label inference attacks in FL are simple, but cannot reveal enough information about a target client because they have largely overlooked the client’s label distribution, especially in the FedGNNs scenario, leaving a significant vulnerability under-exploited.
% \vspace{-3.pt}
\subsection{Label Distribution Inference Attacks}
In the LDI attack, the adversary aims to reconstruct the complete label distribution of a client’s training dataset. 
{Dai et al.} introduced Decaf, which leverages gradient-change magnitudes in the final linear layer to estimate class proportions~\cite{dai2024decaf}. 
% Their method first identifies null classes and then fits the proportions of remaining classes using regression against auxiliary per-class gradient bases. 
Similarly, {Gu et al.} demonstrated that per-class sample proportions leave distinct traces on output-layer updates \cite{gu2023ldia}. 
% Based on this practice, their approach trains shadow models locally and uses the resulting data to train a CNN-based attack model, which maps sequences of output-layer updates to a client’s underlying label distribution. 
Additionally, {Ramakrishna et al.} designed four practical estimators that transform a client’s last-layer update into an estimate of its label distribution, either through a bias term in the final layer or by using an auxiliary dataset \cite{ramakrishna2022inferring}. 

However, \textit{the above methods are only tested on image datasets, which possess a very different structure and data structure from the graph domain}. 
Launching LDI in FedGNNs is challenging because gradients are influenced not only by node labels but also by neighboring structures and unknown node degrees, making direct reverse engineering non-trivial.
Recently, a concurrent work from Cheng et al. proposed EC-LDA, which clips the global model’s parameter norm before distributing it to clients, generates Gaussian-sampled dummy data, and collects embedding summaries with softmax outputs \cite{cheng2025ec}. 
% By combining clipped-model gradients with dummy-data statistics, the attacker reconstructs label counts, reverting to the original model afterward to avoid training degradation. 
% To date, EC-LDA is the only existing work that implements an LDI attack in the FedGNNs setting. 
The success of EC-LDA relies on embedding compression, meaning the server clips the parameters of the global model to minimize the gradient variance before broadcasting it to the clients. 
This method made an aggressive threat model, where the server (attacker) can actively tamper with the client model parameters during the training process.
Such an active modification of the training process makes the attack more detectable and potentially disrupts model utility, which is classified as an active attacker. 
In contrast, our Fed-Listing resides in a passive attacker setting, where the honest-but-curious server only leverages an auxiliary dataset and the attack is undetectable.

\begin{figure*}[t]
  \centering
  \includegraphics[width=\textwidth]{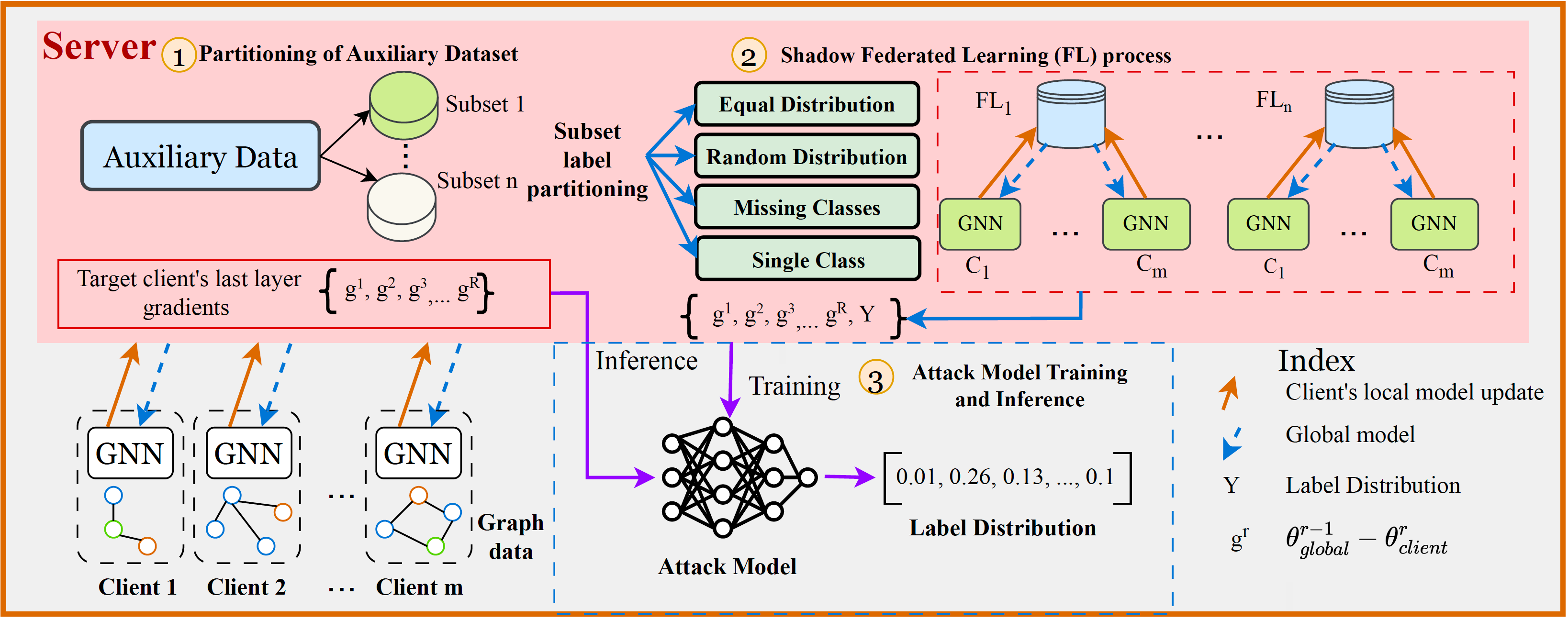}
  \caption{Overview of the proposed attack: Fed-Listing. 
  % All clients train an identical GNN (chosen from three variants) to train on one of four graph datasets. 
  % Each auxiliary data subset is further partitioned based on four criteria: single class, random distribution, equal distribution, and missing classes. 
  % Multiple independent shadow training sessions in the FL setting are performed to generate attack data. 
  } 
  \label{fig:schema}
\end{figure*}

\section{Preliminary}
\label{sec:pre}
\subsection{Graph Neural Networks}
GNN models are neural network-based models designed to efficiently represent graph-structured data. 
A graph data is denoted as $G = (V,E,X)$, where $V$ is the set of $N$ nodes, $E$ is the set of edges, and $X \in \mathcal{R}^{N \times d}$ represents the d-dimensional node feature matrix. 
% The edges are represented by an adjacency matrix $A$ of the graph denoted by $A \in \mathcal{R}^{N \times N}$, where $A_{ij} = 1$ , if there exists an edge between nodes i and j, and $0$ otherwise. 
The fundamental principle of GNNs is message passing from the neighboring nodes, where each node iteratively aggregates information from its neighbors to update its representation. For a node $v$, the representation at the $l^{th}$ layer is given by:
$$h_v^{(\ell)} = \sigma \!\left( 
    W^{(\ell)} \cdot 
    \text{AGG}\!\left( 
        \{ h_u^{(\ell-1)} : u \in \mathcal{N}(v) \} 
        \cup \{ h_v^{(\ell-1)} \} 
    \right) 
\right),
$$
where $h_v^{(\ell)}$ denotes the embedding of node $v$ at layer $\ell$, 
$\mathcal{N}(v)$ represents the neighbors of $v$, 
$W^{(\ell)}$ is a learnable weight matrix, 
$\sigma$ is a nonlinear activation function, 
and \text{AGG} is a permutation-invariant function such as mean, sum, or max. 

\section{Methodology: Fed-Listing}
\label{sec:fed-listing}
In this section, we first discuss the problem setting of the targeted FedGNNs scenario and then detail the attack method and algorithms.

\subsection{Problem Formulation}
Consider the horizontal FedGNNs scenario with a set of $M$ clients: $S = \{S_1, ..., S_M\}$, where each client $S_k$ holds a local graph: $G^{(k)} = (A^{(k)}, X^{(k)})$, with $A^{(k)} \in \mathbb{R}^{N_k \times N_k}$ representing adjacency matrix of edges on $N_k$ nodes and $X^{(k)} \in \mathbb{R}^{N_k \times d}$ is the node-feature matrix in a shared d-dimensional space. 
During the FL training process, all labels of a client $S_k$, denoted as $\mathbf{y}^{(k)} \in \{1,\dots,T\}^{N_k}$, remain on the clients as privacy information. 
The goal of the attacker is to infer the label distribution in the victim client, i.e., per-class proportions.

In FedGNNs, a global graph model, such as Graph Convolutional Network (GCN), depending on the system configuration $h(\,\cdot\,;\,\theta)\colon (A,X)\;\mapsto\;\widehat{Y}\;\in\;\mathbb{R}^{N\times T}$ is initialized at the server with parameter $\theta$. 
At communication round $t$, the server broadcasts the aggregated model from the last round, $\theta^{\,t-1}$, to every participating client in the FL system. 
Each client $S_k$ then performs $E$ local epochs of the local training algorithm on its local private data following Eq.~\eqref{eq:localup}: 
\begin{equation}
    \label{eq:localup}
\theta_k^{\,t}\;=\;\theta^{\,t-1}\;-\;\eta\,\nabla_{\theta}\Biggl[\frac{1}{N_k}\,
\sum_{u=1}^{N_k}
\ell\bigl(h(A^{(k)},X^{(k)})_u;\,y^{(k)}_u\bigr)
\Biggr],
\end{equation}
where $\ell$ is the per‐node cross‐entropy loss and $h(\cdot)_u$ denotes the output (logits) at node $u$. Once local updates are complete, each client sends $\theta_k^t$ back to the server, which aggregates them via weighted averaging, e.g., Federated Averaging (FedAvg)~\cite{fedavg}, 
$\theta^{\,t}
\;=\;
\sum_{k=1}^{M}
\frac{N_k}{\sum_{j=1}^{M} N_j}\,
\theta_k^{\,t}$. 
Equivalently, horizontal FL on graph data seeks to solve the global objective as follows, 
\begin{equation}
\label{eq:agg}
   \theta^*\;=\;\arg\min_{\theta}\sum_{k=1}^M\frac{N_k}{\sum_{j=1}^M N_j}\,\mathbb{E}_{u\sim U_k}\!\Bigl[\ell\bigl(h(A^{(k)},X^{(k)})_u;\,y^{(k)}_u\bigr)\Bigr]. 
\end{equation}

The FedAvg procedure integrates these local optimization steps with server‐side aggregation, enabling the shared GNN parameters to reflect patterns across all sets of disjoint nodes, without exchanging raw adjacency or feature data. 
Formally, the attacker in this setting aims to infer the label distribution of the target client $S_k$ as denoted by $\{y^{(k)}_1, y^{(k)}_2, \dots, y^{(k)}_T\}$, where $y_t^{(k)}$ denotes proportion of class $c$ and $\sum_{t=1}^T y^{(k)}_t=1$.

\subsection{Threat Model}
In our attack setting, the FL server acts as an honest-but-curious adversary, attempting to infer the label distribution of a target client's training data. 
While the FL process itself remains unaffected, the server can exploit the last-layer gradients of the target client on each FL round to carry out the inference. 

\noindent\textbf{Adversary's knowledge}: As the server serves as the attacker, it has full access to the parameters shared by the clients and the label space of the training dataset. 
Also, the server has access to an auxiliary dataset, which consists of a similar distribution and the same label space as the real dataset of FL clients. 

\noindent\textbf{Adversary's capability}: The adversary can leverage the model parameters received from each client across each round to extract information and deduce the underlying label distribution of the target client's local data. 

\subsection{Fed-Listing Schema}

% In this section, we provide a detailed description of the Fed-Listing method. 
% The primary method involves dividing the Auxiliary dataset into $n$ subsets, with each subset consisting of approximately equal proportions of data samples of each class. 
% After that, the dataset in each subset is further split into different clients to generate the attack data with simulated FL training. 
% The partition of the auxiliary subset data into various clients is done based on the four categories: single class, random distribution, equal distribution, and missing classes. 
% For each category, an independent FL is trained with each client having its local data class distribution.  
% In all the FL rounds, the gradient of the last layer of the model is recorded for all the clients. Once the training concludes, these recorded gradients, with their corresponding labels, are used as the training dataset. For instance, if there are 10 clients and FL training occurs for 50 rounds, then we obtain 10 feature vectors of size 50. 
% In all the FL rounds, the gradient is collected, which is denoted by: 
% $G = [g_1, g_2, g_3, ..., g_R],$ where $ g_R = \theta_{global}^{R-1} - \theta_{client}^R$. 
% This dataset is used to train a neural network-based attack model. Since we are trying to bring the ground truth label distribution and the predicted label distribution as close as possible, we used a combination of mean squared error, mean error, and JS-div error with a scalar added to it. 
% $$ L = a * L_1 + b* L_2 + c * L_3$$

In this section, we provide a detailed description of the proposed Fed-Listing method. 
The attack proceeds in three main phases: 
(1) partitioning of the auxiliary dataset, 
(2) {shadow FL training process}, 
and (3) attack model training and inference as shown in Figure \ref{fig:schema}. 

\subsubsection{Partitioning of the Auxiliary Dataset}
The auxiliary dataset $\mathcal{D}_a$ was first divided into $n$ subsets, each containing approximately equal proportions of samples across all classes. 
Each subset was further partitioned into multiple clients to simulate FL participants. 
To capture diverse local data distributions, we design four partitioning strategies (or scenarios): 
(i) Equal proportion: all classes are equally represented in each client, 
(ii) Random distribution: classes are assigned randomly across clients, reflecting the non-i.i.d. case, 
(iii) Single-class: each client is assigned samples from only one class, 
and (iv) Missing classes: some classes are absent from certain clients.
% and (v) One-class dominant: one class has a much higher proportion compared to the others.
These partitions were selected to cover the scenarios that are i.i.d., non-i.i.d., and extremely heterogeneous in FL~\cite{zhao2018federated}. 

% \textbf{For each partition category, an independent FL instance was conducted to ensure that local data heterogeneity was fully represented. }
Since auxiliary datasets are typically limited in size, resampling techniques across different partitioning categories were permitted. 
% However, within any given FL scenario, the data assigned to each client remained independent to preserve the integrity of the local distributions. 
For each scenario, we empirically report the number of FL instances and the number of clients with a given (non-i.i.d.) label distribution required for the attack to succeed, and the results are shown in the supplementary material. 
For example, in the \textit{Random distribution} setting, we observe that the attack performed better as we increased the number of FL instances. 
% {\color{red} simulation number. what is simulation number? This statement does not match context.}
Similarly, for \textit{Single-class} and \textit{Missing-class} settings, we show that the FL instance with clients having a higher proportion of each case yielded higher attack performance. 

\subsubsection{Shadow FL Training Processes}
During federated training, in each communication round $r$, the gradients of the last layer are collected from all participating clients.
Specifically, the gradient update for round $r$ is defined as follows,
\begin{equation}
\label{eqn:grad}
g^r = \theta^{r-1}_{global} - \theta^r_{client}, \in \mathbb{R}^{d_{out}},
\end{equation}
where $\theta^{r-1}_{global}$ is the global model distributed at the beginning of round $r$, $\theta^r_{client}$ is the locally updated model parameter after round $r$, and $d_{out}$ is the output dimension. 
Over $R$ communication rounds, the complete gradient record for a client $k$ is represented as,
$$ G_k = [g^1_k, g^2_k, ..., g^R_k] \in \mathbb{R}^{R \times{d_{out}}}, $$
% For example, if there are $C$ clients and $R$ communication rounds, the server can obtain $C$ feature vectors of size $R$ each. 
For each data partitioning setting, we collected the last-layer gradients from individual clients, yielding a set of gradient records 
$\{G_k\}_{k=1}^i$, where $i$ denotes the number of clients. 
Each client’s gradients have their corresponding label distribution, which serves as the ground-truth target for training. 
To construct the attack dataset, the gradients from all clients are concatenated and flattened into fixed-length feature vectors to ensure compatibility with the neural network-based attack model. 
Exploiting the parameters or gradients of the final layer is a popular approach for inferring training data, specifically for recovering labels and label distributions, as demonstrated in several prior works~\cite{gu2023ldia, dai2024decaf, wang2024breaking}.

\subsubsection{Attack Model Training and Inference}

\begin{algorithm}[tb]
\caption{Fed-Listing Algorithm}
\label{alg:fedlisting}
\begin{algorithmic}[1]
\Statex \textbf{Input:} Auxiliary dataset $\mathcal{D}_a$, clients $C$, rounds $R$, local epochs $E$, partition strategies $\mathcal{P}$, loss weight $\alpha$
\Statex \textbf{Output:} Trained attack model $\hat{\mathcal{A}}$
\newline Split $\mathcal{D}_a$ into subsets $\{\mathcal{D}_a^{(1)},\dots,\mathcal{D}_a^{(n_s)}\}$
\For{each subset $\mathcal{D}_a^{(s)}$ and strategy $p \in \mathcal{P}$}
  \State Partition $\mathcal{D}_a^{(s)}$ into $C$ clients
  \For{$r=1$ to $R$}
    \For{each client $k$}
      \State Train locally for $E$ epochs
      \State Record last-layer update $g_k^{r}=\theta^{r-1}_{\text{global, last layer}}-\theta^{r}_{k \text{, last layer}}$
    \EndFor
    \State Aggregate updates $\to \theta^{r}_{\text{global}}$
  \EndFor
  \For{each client $k$}
    \State $G_k \gets [g_k^{1},\dots,g_k^{R}]$, flatten to $G_k^{\text{flat}}$
    \State $Y_k \gets$ label distribution of $D_k$
    \State Add $(G_k^{\text{flat}}, Y_k)$ to attack dataset $\mathcal{T}_{attack}$
  \EndFor
\EndFor
\State Train attack model $\mathcal{A}$ on $\mathcal{T}_{attack}$ with loss
\[
   \mathcal{L}(Y,\hat Y) = \alpha L_{L1}(Y,\hat Y) + (1-\alpha)L_{JS}(Y,\hat Y) ,
\]
\State \Return trained attack model $\hat{\mathcal{A}}$
\end{algorithmic}
\end{algorithm}

The recorded gradients, paired with their corresponding label distributions, form the training dataset for the attack model. 
Formally, each training sample is represented as $(G_k,Y)$, where $Y = \{y_1, y_2, \dots, y_T\}$ is the ground truth label distribution of the client, with $\sum_{i=1}^T y_i = 1$.
Then, a neural network–based attack model is trained to infer label distributions of clients. 
The objective is to minimize the discrepancy between the predicted label distribution $\hat{Y}$ and the ground truth $Y$. 
To achieve this, we employ a composite loss function that combines multiple distribution-similarity measures:

\begin{equation}
    \mathcal{L}(Y,\hat Y) = \alpha L_{L1}(Y,\hat Y) + (1-\alpha) L_{JS}(Y,\hat Y) ,
    \label{eqn:loss}
\end{equation}
where $L_{L1}(Y,\hat{Y}) = \frac{1}{T} \sum_{t=1}^{T} \left| y_t - \hat{y}_t \right|$ is the $L_1$ norm
and 
$ L_{JS}(Y \,\|\, \hat{Y}) 
= \frac{1}{2}\, \text{KL}(Y \,\|\, M) 
+ \frac{1}{2}\, \text{KL}(\hat{Y} \,\|\, M),$
is the Jensen-Shannon divergence (JS-divergence).
The parameter $\alpha$ is a scalar weight balancing the contribution of each term. 
We chose the two types of loss functions for the following reasons: 
The alignment loss (i.e., L1 norm) compares individual elements in two distributions and is generally used to measure the statistical discrepancy between two data distributions. 
The distribution matching loss (i.e., JS-divergence) addresses the degree of overlap or separation between the distinct distributions. 
% We empirically selected the objective function(s) that yielded the best performance as shown in Table \ref{tab:loss_ablation} and fixed the loss function form in Eq.~\eqref{eqn:loss}.

The optimum values for hyperparameters $\alpha$ were selected using grid search on values from $0$ to $1$ with a step size of $0.05$ (shown in Figure~\ref{fig:alpha_vs_perf}). 
The value that resulted in the highest performance was selected as the optimum value.
For the Citeseer and Amazon computer datasets, the best performance was observed when $\alpha$ was set to $0.05$, whereas PubMed and Cora had the best result when the value of $\alpha$ was set to $0.00$.

\section{Experiment Results}
\label{sec:exp}
In this section, we simulate the proposed attack method, Fed-Listing, on real-life datasets, compare it with the existing attack baselines, and evaluate the attack's robustness against popular defense mechanisms.

\subsection{Experiment Settings}

\textbf{Datasets.}
To evaluate the effectiveness of Fed-Listing, we conducted experiments on four widely used graph benchmarks~\cite{sen2008collective,shchur2018pitfalls}: Cora, PubMed, Citeseer, and Amazon Computers, under varying client label distributions. 
These datasets are chosen to assess the accuracy of our proposed attack in recovering each client’s label distribution.
Cora, PubMed, and Citeseer are citation network datasets, where the nodes correspond to scientific papers and the edges represent citation links. 
% In Cora and Citeseer, each paper is encoded using a Bag-of-Words (BoW) feature vector, while in PubMed, features are represented using Term Frequency–Inverse Document Frequency (TF-IDF) weighted word vectors. 
Amazon Computers is a co-purchase network, where the nodes represent products in the “Computers” category, and the edges denote co-purchase relationships -- such as laptops purchased alongside covers or mice. 
% Product features in this dataset are also represented using BoW vectors derived from metadata, e.g., product descriptions and reviews.
A detailed summary of the datasets and their properties is provided in Table~\ref{tab:data_info}.

\begin{table}[t]
    \centering
    \small
    \caption{Description of datasets including the number of nodes, edges, features, and label classes on each dataset.}
    \begin{tabular}{ccccc}
    \hline
    \textbf{Dataset} & \textbf{Nodes} & \textbf{Edges} & \textbf{Features} & \textbf{Classes}\\
    \hline
      Cora & 2,708 & 5,429 & 1,433 &  7  \\
    PubMed  & 19,717 & 44,338 & 500 & 3 \\
    Citeseer & 3,327 & 4,732 & 3,703 & 6 \\
    Amazon Computers & 13,752 & 245,861 & 767 & 10 \\
    % Amazon Photo & 7,487 & 119,043 & 745 & 8 \\
    \hline
    \end{tabular}
    
    \label{tab:data_info}
\end{table}

\textbf{GNN Architecture and Baselines.}
We evaluate our method using three widely adopted GNN architectures: GCN~\cite{kipf2016semi}, GraphSage~\cite{hamilton2017inductive}, and Graph Isomorphism Network (GIN)~\cite{xu2018powerful}. 
% While all these models follow the fundamental message-passing paradigm, where node representations are updated by aggregating information from neighboring nodes, they differ primarily in their aggregation strategies.
GCN generalizes convolution operations to graphs by applying a linear transformation to node features aggregated via a normalized adjacency matrix. 
GraphSage improves scalability by sampling a fixed-size set of neighbors and aggregating their features using a learnable function. 
% GAT introduces an attention mechanism that dynamically weights neighbors during aggregation, enabling the model to assign greater influence to more informative neighbors. 
GIN, recognized for its expressive power, employs a sum aggregator followed by a Multilayer Perceptron (MLP), allowing it to capture complex graph structures with high fidelity.

To assess the effectiveness of our Fed-Listing attack, we compare it against three baselines. 
The first one is random guessing, which estimates the label distribution of each target client by sampling from a normal distribution, providing a minimal performance reference. 
The second baseline is Decaf~\cite{dai2024decaf}, a recent method designed for LDI attack in FL, but on image data.
We configure the source code to make it fit the graph data input as a passive attacker baseline. 
We include Decaf as it shares a similar threat model and evaluation setting, particularly in scenarios where clients exhibit uniform or skewed label distributions, making it a strong and relevant baseline for comparison.
The third baseline is EC-LDA~\cite{cheng2025ec}, which is designed as an active attacker by manipulating the training parameters and FL process. 
Though the active attacker has the limitation of being detected, it can serve as a practical upper bound, justifying the performance of our passive attack methods.

\textbf{Evaluation Metrics.}
Since the attacker infers the proportion of each label, we compare the alignment of the inferred distribution $\hat{Y} = (\hat{y}_1, \dots, \hat{y}_T)$ and the ground truth label distribution $Y=(y_1, \dots, y_T)$. 
For consistency in previous works~\cite{cheng2025ec, gu2023ldia}, we selected three evaluation metrics: Manhattan Distance (MD, also known as L1 norm), JS-divergence, and Cosine Similarity (CS). 
JS-divergence is symmetric and bounded between 0 and 1, making it a stable metric to measure the distributional divergence. 
In contrast, CS measures the directional similarity of the two distributions and shows how well the predicted and true distributions align in the same direction. 
For MD and JS-divergence, smaller values indicate a better match to the ground truth distribution. 
Whereas higher values of the CS suggest stronger alignment with the ground truth. 

\textbf{Attack Scenarios.}
An auxiliary dataset typically consists of samples drawn from a similar (or a closely related) distribution and is employed to train shadow models for performing attacks. 
We partition each dataset $\mathcal{D}$ into two \textbf{disjoint} subsets: the training set $\mathcal{D}_T$ and the auxiliary set $\mathcal{D}_a$. 
In this partitioning, we reserve $20\%$ of the original data as $\mathcal{D}_a$ for our attack to succeed.
% {\color{red} auxiliary dataset? What kind of auxiliary data is needed?}
To evaluate the effectiveness of the attack, we considered five scenarios with varying class distribution proportions, as summarized in Table~\ref{tab:result_gcn}. 
% \ref{tab:result_sage}, \ref{tab:result_gin}. 

The auxiliary dataset may originate from the same distribution and share an identical label space with the target task. 
Auxiliary data drawn from a different domain can also be used, provided that it shares the same label space as the target dataset. 
In cases where the label spaces do not fully align, or data is not publicly available, a synthetic class (or data) can be introduced to initiate the attack. 
To illustrate a real-world scenario, we evaluated the attack performance on the synthetic auxiliary data as shown in Table~\ref{tab:result_gcn}. 
The synthetic graph was generated using GraphUniverse~\cite{van2025graphuniverse}, which produces a graph family (i.e., homophily graphs) at scale, aligning with the data we used for our experiments.

\textbf{Defense Strategies.}
We evaluate the robustness of Fed-Listing under three widely adopted defense strategies: Differential Privacy (DP)~\cite{abadi2016deep}, noisy gradients~\cite{zhu2019deep}, and gradient compression~\cite{lin2017deep}. 
DP and noisy gradient apply a noisy perturbation to the training process of FedGNNs. 
% ensures that the influence of any single training example on the shared model update is bounded, typically enforced through the $(\epsilon, \delta)$-DP, where $\epsilon$ controls privacy loss and $\delta$ denotes the probability of failure. 
% Using the Gaussian mechanism, noise with standard deviation $\sigma = \frac{\sqrt{2 \ln(1.25/\delta)}}{\epsilon}$ is added to clipped gradients, balancing privacy and utility. 
% The noisy gradient defense applies noisy perturbation directly to client gradients in each communication round, with the update $\tilde{g} = g + \mathcal{N}(0, \sigma^{2}I)$, where $g$ denotes the original gradient and $\sigma$ scales the injected noise.
% This additional noise reduces information leakage while maintaining training feasibility. 
Gradient compression reduces the communication footprint and obscures fine-grained information by transmitting only the significant gradient elements with the largest magnitudes, defined as $\tilde{g}_i = g_i$ if $|g_{i}| \geq \tau$, and $\tilde{g}_{i} = 0$ otherwise, where $\tau$ is the threshold chosen such that only a fraction $\alpha$ of gradients are preserved. 
Together, these defenses aim to diminish the attack surface of inference attacks in FL.
\begin{table*}[tbh]
  \centering
  \caption{Performance under different client‐data splits and distance/divergence measures. All reported metrics were obtained using the GCN model in both FL training and shadow FL training.  
  The highest performance is denoted in \textbf{bold} text. A: Auxiliary data, S: Synthetic auxiliary data, AC: Amazon Computers.}
\label{tab:result_gcn}
  \resizebox{\textwidth}{!}{%
    \begin{tabular}{@{} ll ccccc ccccc ccccc @{}}
      \toprule
      \textbf{Datasets}      & \textbf{Target label distribution}       
                    & \multicolumn{5}{c}{\textbf{Manhattan distance ($\downarrow$)} }
                    & \multicolumn{5}{c}{\textbf{Jensen-Shannon divergence ($\downarrow$)} }
                    & \multicolumn{5}{c}{\textbf{Cosine similarity ($\uparrow$)}} \\
      \cmidrule(lr){3-7} \cmidrule(lr){8-12} \cmidrule(lr){13-17}
                   &                                
                    & Ours (A) & Ours (S)  & Random & Decaf & EC-LDA 
                    & Ours (A) & Ours (S) & Random & Decaf & EC-LDA
                    & Ours (A) & Ours (S) & Random & Decaf & EC-LDA  \\
      \midrule
      \multirow{5}{*}{Cora}
        & Equal proportion                           
                    & \textbf{0.221} & 0.562   & 0.399    &  0.422 &  {0.000}
                  &  \textbf{0.001}  & 0.009 & 0.003    &  0.004 &  {0.000} 
                  &   \textbf{0.973} & 0.819 &  0.918   &  0.871 &  {1.000} \\
        & Random split    
                    & \textbf{0.183}  & 0.308  & 0.471    &  0.796 &  {0.012}
                  &  \textbf{0.002} & 0.006 & 0.007    &  0.025 &   {0.000}
                  &   \textbf{0.971} & 0.884 &  0.888   &  0.547 &  {0.999} \\
        & One class missing   
                    & \textbf{0.183}  & 0.637 & 0.471    &  0.635 &  {0.006}
                  &  \textbf{0.002} & 0.017 & 0.007    &  0.119 & 0.005
                  &   \textbf{0.971} & 0.765 &  0.888   &  0.762 &  {1.000} \\
        & Single class only                      
                    & 1.287  & \textbf{1.168} & 1.787    &  1.350 &  {0.000}
                  &  \textbf{0.035} & 0.316 & 0.053    &  0.375 &   {0.000}
                  &   0.778 & \textbf{0.864} &  0.223   &  0.734 &  {1.000} \\
        & One-class dominant                       
                     & 1.095 &  1.389  & \textbf{0.968}    &  1.335 &  {0.000}
                  &  \textbf{0.029} & 0.058  & 0.033    &  0.062 &  {0.000} 
                  &   0.704 & 0.266  &  \textbf{0.766}   &  0.399 &  {1.000} \\
      \midrule
      \multirow{5}{*}{PubMed}
         & Equal proportion                         
                   & \textbf{0.166} &  0.271 & 0.315    &  0.336 &  {0.000}
                  &  \textbf{0.001} & 0.004 & 0.005    &  0.006 &  {0.000}
                  &  \textbf{0.982} & 0.960  &  0.948   &  0.941 &  {1.000} \\        
        & Random split  
                  & \textbf{0.093}  & 0.286 & 0.582    &  0.812 &  {0.012}
                  &  \textbf{0.000} & 0.004 & 0.017    &  0.559 & 0.000
                  &   \textbf{0.994} & 0.956  &  0.821   &  0.698 &  {0.999} \\    
        & One class missing    
                  & \textbf{0.243} & 0.490  & 0.382    &  0.353 &  {0.059}
                  &  \textbf{0.003} & 0.323  & 0.007    &  0.008 & 0.010
                  &   \textbf{0.968} & 0.895  &  0.918   &  0.913 &  {0.999}  \\  
        & Single class only                       
                   & 1.196  & 1.312 & \textbf{0.940}    &  1.000 &  {0.000}
                  &  \textbf{0.081} & 0.901   & 0.620    &  0.370 &  {0.000}
                  &   0.683 & 0.594 &  \textbf{0.842}   &  0.520 &  {1.000} \\  
        & One-class dominant                      
                & 0.783 & \textbf{0.546}  & 1.583    &  0.708 &  {0.013}
                  &  0.030 & \textbf{0.017} & 0.275    &  0.037 &  {0.000}  
                  &   0.812 & \textbf{0.861} &  0.188   &  0.795 &  {1.000} \\  
      \midrule
      \multirow{5}{*}{Citeseer}
         & Equal proportion                          
                  & 0.152 & 0.451  & 0.334    &  \textbf{0.147} &  {0.000}
                  &  \textbf{0.000} & 0.007 & 0.002    &  0.001 & 0.000
                  &   \textbf{0.985} & 0.864 &  0.943   &  0.980 &  {1.000} \\  
        & Random split    
                    & \textbf{0.107} & 0.317  & 0.336    &  0.244 &  {0.008}
                  &  \textbf{0.000} & 0.002 & 0.004    &  0.002 &  0.000
                  &   \textbf{0.993} & 0.937 &  0.918   &  0.956 &  {0.999} \\  
        & One class missing      
                  & \textbf{0.412} &  0.364  & 0.625    &  0.692 &  {0.017}
                  &  \textbf{0.011}& 0.045  & 0.115    &  0.115 &  {0.002}
                  &   \textbf{0.902}& 0.916  &  0.800   &  0.783 &  {0.999} \\  
        & Single class only                          
                   & 1.676 & 1.614   & 1.615    &  \textbf{1.292} &  {0.003}
                  &  0.568&  0.542 & 0.544    &  \textbf{0.050} &  {0.000}
                  &   0.391 & 0.470 &  0.450   &  \textbf{0.469} &  {1.000}\\  
        & One-class dominant                       
                     & \textbf{0.928}& 1.288   & 1.467    &  1.217 &  {0.029}
                  &  \textbf{0.039} & 0.047 & 0.082    &  0.042 &  {0.002}
                  &   0.559 & 0.420  &  0.304   &  \textbf{0.710} &  {0.999} \\  
      \midrule
      \multirow{5}{*}{AC}
          & Equal proportion         
                  & \textbf{0.435} &  0.616 & 0.489    &  0.706 &  {0.000}
                  &  \textbf{0.003}& 0.007  & 0.004    &  0.084 &  {0.000}
                  &   \textbf{0.903}& 0.823 &  0.852   &  0.746 &  {1.000} \\
        & Random split    
                &  \textbf{0.225} & 0.713  &  1.050   &  0.953 &  {0.089}
                  &   \textbf{0.001}& 0.012  &  0.023   &  0.016 &   {0.002}
                  &   \textbf{0.966}& 0.733 &  0.493   &  0.564 &  {0.998} \\
        & One class missing    
                   &  \textbf{0.617} & 0.806 &  0.805    &  0.665 &  {0.000}
                  &  0.068 & 0.061 &  0.034   &  \textbf{0.027} &  {0.000}
                  &  \textbf{0.858} &  0.704 &  0.068   &  0.810 &  {1.000} \\
        & Single class only                        
                    &  \textbf{0.860}& 1.810  &  1.732   &  1.874 &  {0.027}
                  &  \textbf{0.038} & 0.367 &  0.351   &  0.387 &  {0.004}
                  &   \textbf{0.771} & 0.272 &   0.344  &  0.189 &  {1.000} \\
        & One-class dominant                     
                  &  \textbf{1.415}  & 1.429 &  1.543    &  1.544 &  {0.081}
                  &   \textbf{0.043} & 0.044  &  0.069   &  0.063 &  {0.003}
                  &  \textbf{0.361} & 0.325 &  0.069   &  0.166 &  {0.999} \\

      \bottomrule
    \end{tabular}%
    
  }  
\end{table*}

\textbf{Implementation Details.}
We implemented Fed-Listing using the Flower FL framework (v1.19.0), with integration of PyTorch (v2.7.1) and PyTorch Geometric (v2.6.1) for GNN training. 
Each experiment employed a consistent two-layer GNN architecture across all clients to ensure comparability. 
We used $C=10$ clients and conducted $R = 50$ communication rounds. 
For local training, each client applied the Adam optimizer with a learning rate of 0.001 and a batch size of 32. 
To support the attack, we recorded the gradients of the final GNN layer from each client at the end of every communication round. 
These gradients were then transformed into structured feature vectors, serving as inputs to our attack model.

The attack model was designed as an MLP with two hidden layers containing 256 and 128 units, respectively, and ReLU activations. 
% The training was performed on last-layer gradients obtained from shadow datasets created via four different client partitioning strategies: single-class, random distribution, equal distribution, and missing-class configurations. 
In experiments with limited auxiliary data, we allowed repeated use of samples across different partitioning strategies while ensuring that each client’s data remained independent within a given federated setting. 
% The attack model was optimized using a composite loss function as defined in Eq.~\eqref{eqn:loss}.

\subsection{Attack Performance and Analysis}
The results in Table~\ref{tab:result_gcn} illustrate the effectiveness of Fed-Listing across all datasets and partitioning strategies, on real and synthetically generated auxiliary data, when using GCN as a local client model. 
Our attack method generally achieves the lowest MD and JS-divergence, as well as the highest CS, compared to both random guessing and Decaf. 
Specifically, under the equal proportion setting, our Fed-Listing performs nearly perfectly, with JS-divergence as 0.000 and CS exceeding 0.985 on the Citeseer dataset. 
Non-i.i.d. scenarios highlight the clear superiority of our approach against the two baselines (Random and Decaf) in GCN.
For example, in the challenging One-class missing and One-class dominant scenario, where clients hold highly skewed data distributions, our Fed-Listing maintains the best performance (e.g., CS of 0.971 and JS-divergence of 0.002 on the Cora dataset), while Decaf and random guessing show a significant decline. 
Similar performance of our Fed-Listing method can be observed on GraphSage and GIN models (shown in the Supplementary material). 
However, for non-i.i.d. cases, the performance gap is narrow, and Decaf sometimes overtakes on specific metrics. 
Notably, Decaf exhibits moderate success in some balanced settings but struggles significantly in highly skewed distributions, particularly evident on Amazon Computers, where its CS drops drastically. 
Additionally, on synthetically generated auxiliary data, our method outperforms both baselines in the random-split scenario across all datasets and generally performs better in non-i.i.d. cases (One-class missing and Single-class only) on Citeseer and Amazon Computers.
The best results occur in the one-class dominant setting on PubMed, with MD, JS-divergence, and CS of 0.546, 0.017, and 0.861, respectively. 
However, performance still lags behind real auxiliary data, likely because feature vectors and edge densities are generated randomly without any optimization.

\textbf{Remark.}
The performance of EC-LDA is displayed for reference in the tables. 
As expected, the active attack EC-LDA, due to its stronger attack assumption that purposely manipulates the model parameters during FL training, allows the malicious server to deceive the local client into revealing more information in the next training round. 
Therefore, the EC-LDA attack can achieve a nearly perfect LDI attack result, yet it comes with the vulnerability of easy detection and the cost of training accuracy for GNN tasks.

Generally, across all methods on the majority of experiments, we observe that the LDI is easier in the Equal and Random split settings, but more challenging in the Single class only and One-class dominant scenarios. 
% This difficulty likely arises because the gradients in these cases were dominated by the majority of clients having more balanced class proportions.
\uline{Due to page limit, the complete experiment results for GraphSage and GIN models of all settings are provided in Tables 1 and 2 in the appendix.}

\begin{figure*}[t]
	\subfigure[Differential Privacy]{
			\centering
			\includegraphics[width=0.3\linewidth]{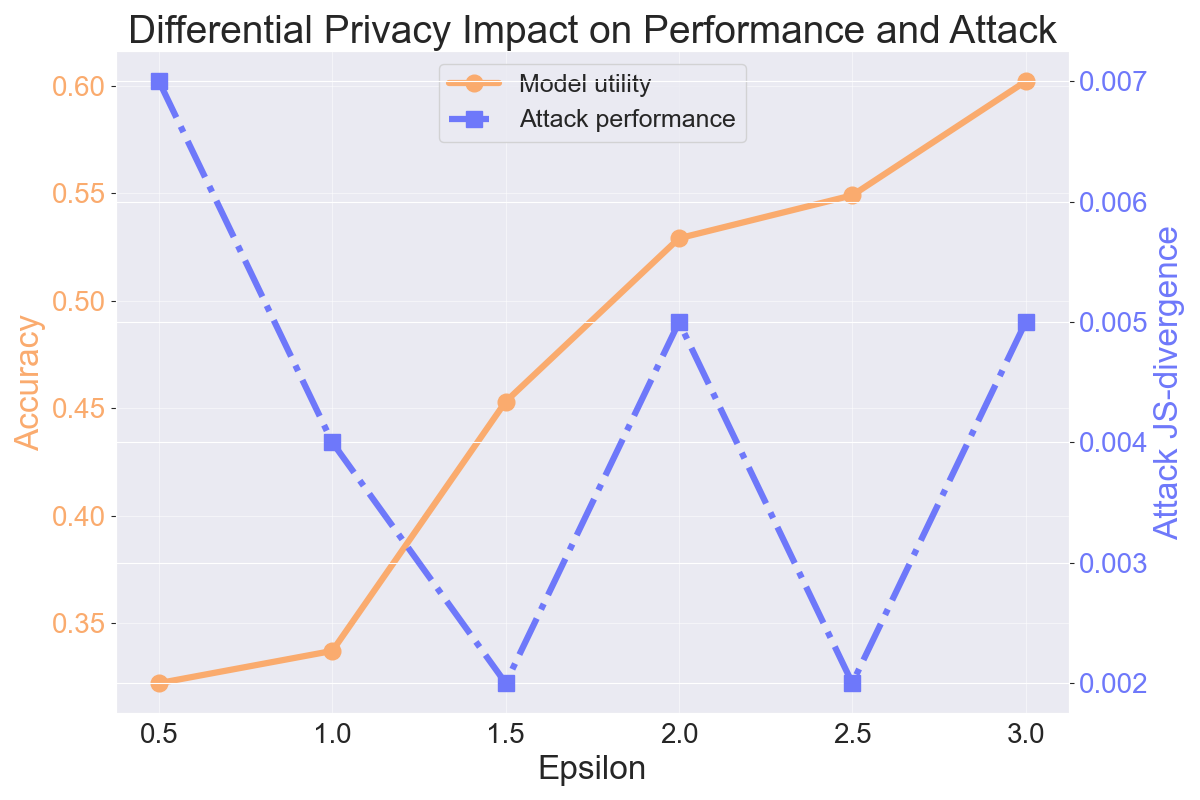}
			\label{fig:def-dp}
	}
	\subfigure[Noisy gradient]{
			\centering
			\includegraphics[width=0.3\linewidth]{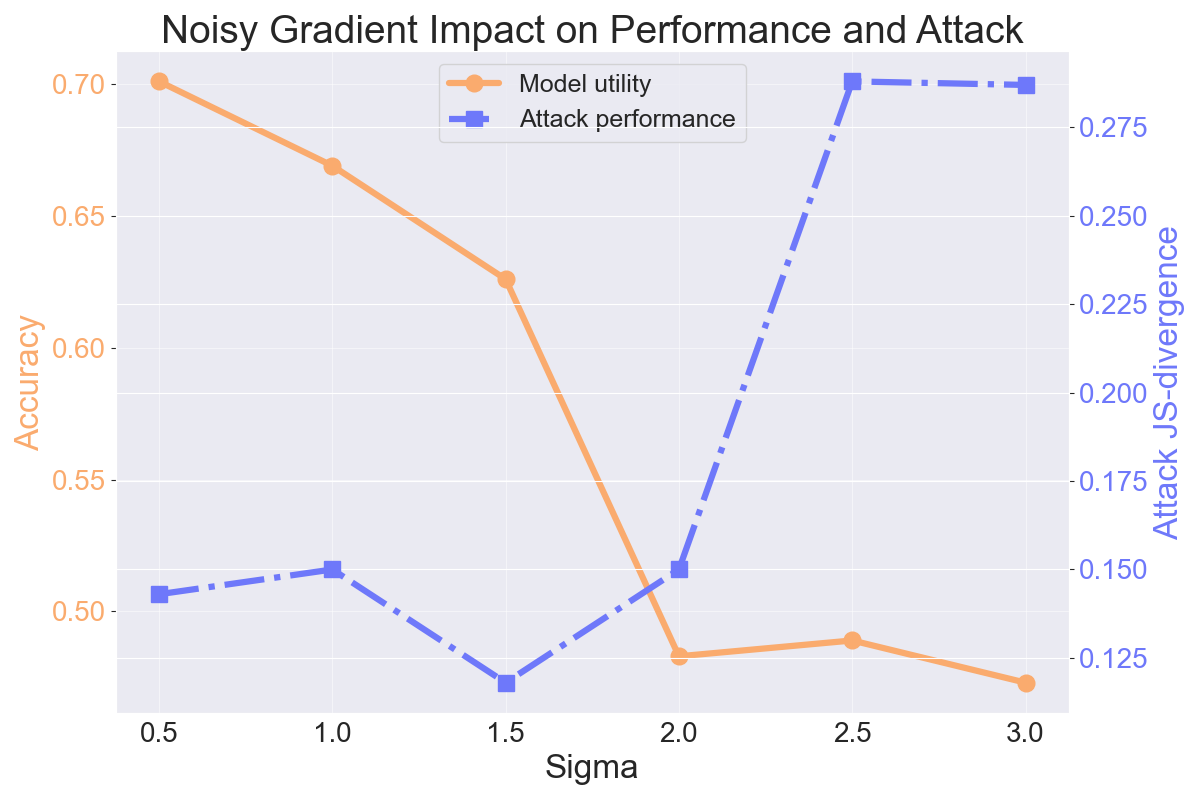}
			\label{fig:def-ng}
	}
    	\subfigure[Gradient compression]{
		
			\centering
			\includegraphics[width=0.3\linewidth]{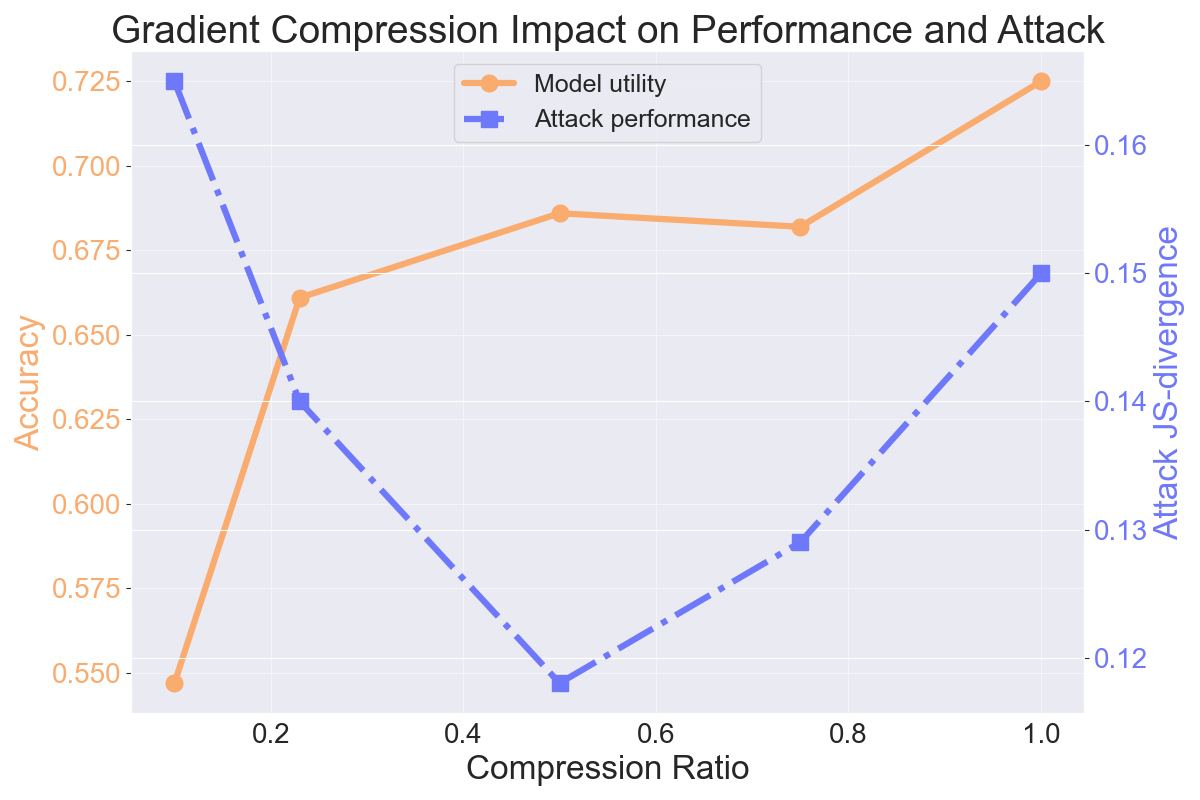}
			\label{fig:def-gc}
	}
	\caption{Model utility (accuracy) and attack effectiveness (JS-divergence) under three defense strategies:  (a) Differential privacy, (b) Noisy gradient, and (c) Gradient compression. 
    We measure the robustness using the GCN model and the Cora dataset.}  
%   \label{fig:defense}
	\label{fig:defense}
\end{figure*}

\subsection{Attack Robustness against Defense Strategy}
% {\color{red} Redraw the figures and rewrite the section.
Figure~\ref{fig:defense} compares the impact of DP, noisy gradients, and gradient compression on the attack robustness of Fed-Listing (in terms of JS-divergence). 
Under the DP defense, the performance of FedGNNs improves as the privacy budget $\epsilon$ increases, as shown in Fig.~\ref{fig:def-dp}.
However, the attack becomes significantly more effective when the privacy budget $\epsilon$ increases, except when $\epsilon$ equals 2 and 3, where the attack performance was reduced.
% Lower $\epsilon$ values reduce the success of the attack but slightly compromise model performance, reflecting the inherent trade-off between utility and privacy. 
In the case of noisy gradients, increasing the noise level $\sigma$ leads to a steep decline in model accuracy from 0.701 at $\sigma = 0.5$ to just 0.473 at $\sigma = 3$. Meanwhile, attack effectiveness remains moderate at intermediate noise levels but deteriorates further at higher noise scales (e.g., $\sigma \geq 2.5$), where both model utility and attack reliability break down.
In contrast, gradient compression demonstrates a more favorable trade-off. 
The accuracy steadily improves with a higher compression ratio, from 0.547 at $\alpha = 0.1$ to 0.725 at $\alpha = 1.0$, while attack success is notably minimized around $\alpha = 0.5$ (JS-divergence = 0.118).

In summary, it is hard for these three defense mechanisms to mitigate the proposed Fed-Listing attack with slight defense strength, yet a stronger defense strength (e.g., larger $\epsilon$ and noise scale $\sigma$) can destroy the model utility.

\section{Ablation study}
\label{sec:ablation}
In this section, we discuss the contribution of various components of the attack model, specifically the impact of the loss function, comparison of performance when exploiting the all or last-layer parameters, and the performance of two different attack models (CNN-based and MLP-based).

\subsection{The Impact of $\alpha$ in Loss Function}

\begin{figure*}[tbh]
	\subfigure[Manhattan distance vs alpha values]{
			\centering
			\includegraphics[width=0.3\linewidth]{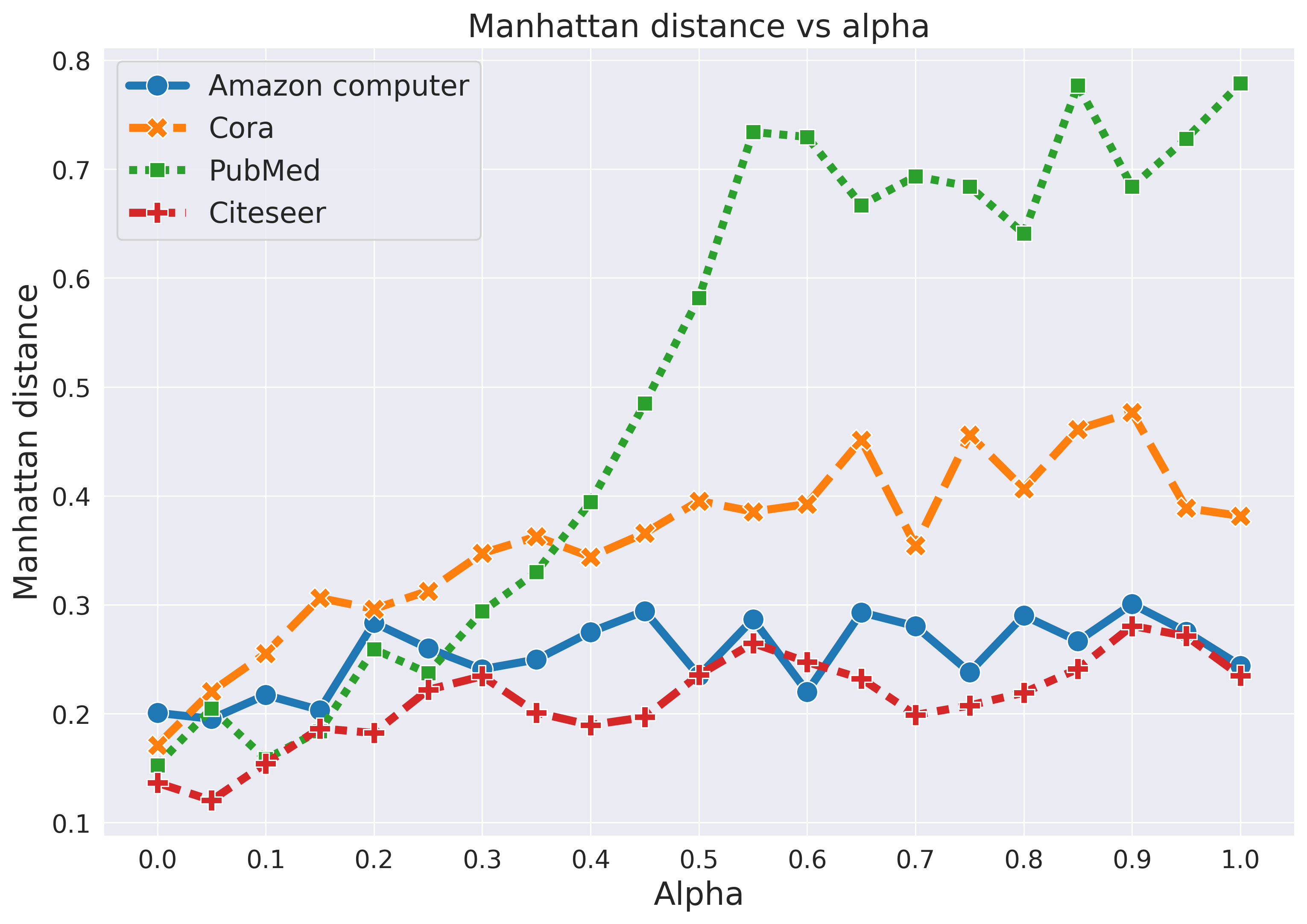}
			\label{fig:p2-client}
	}
	\subfigure[Jensen-Shannon divergence vs alpha values]{
			\centering
			\includegraphics[width=0.3\linewidth]{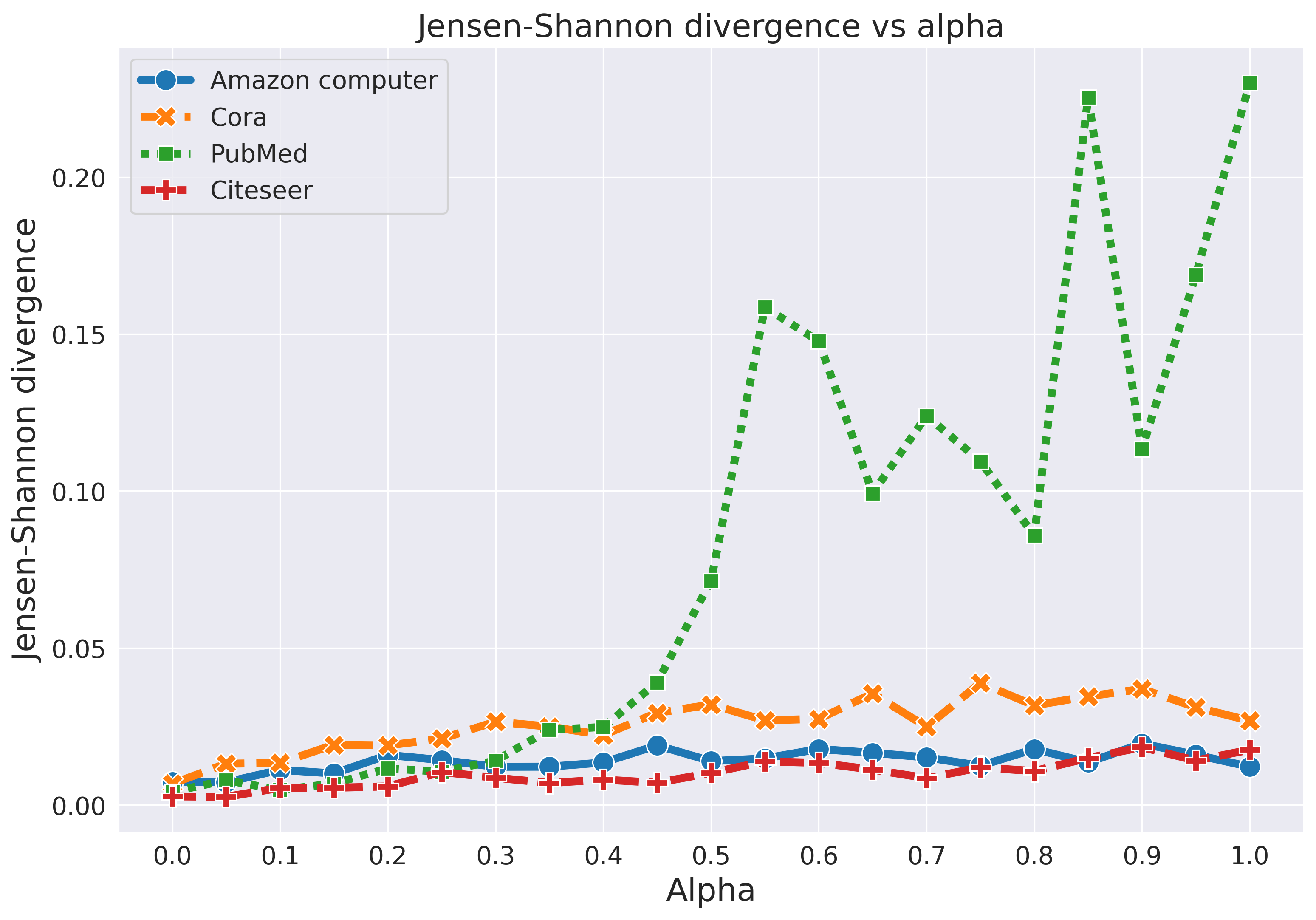} % Used .png file as .pdf was not rendering
			\label{fig:p3-client}
	}
    	\subfigure[Cosine similarity vs alpha values]{
			\centering
			\includegraphics[width=0.3\linewidth]{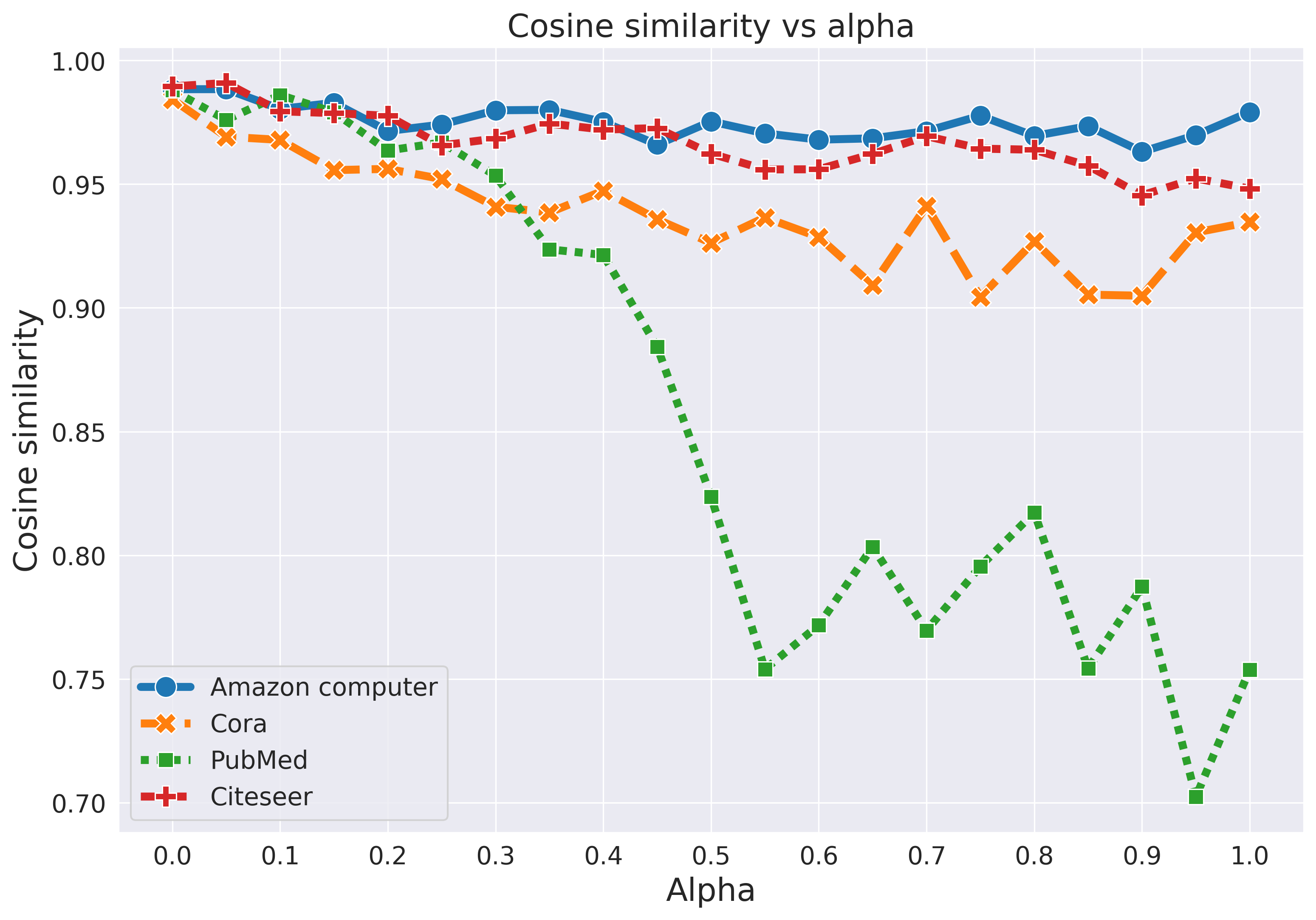} % Used .png file as .pdf was not rendering
			\label{fig:p4-client}
	}
	\caption{This figure depicts the changes in the attack performance when varying the $\alpha$ values.  
    Subfigures (a), (b), and (c) show the Manhattan distance, Jensen-Shannon divergence, and Cosine similarity against different alpha values, respectively. 
    Note that all figures were plotted using GCN as a target model in training both the FL and shadow FL models.}  
	\label{fig:alpha_vs_perf}
\end{figure*}

Figure~\ref{fig:alpha_vs_perf} shows the impact of varying the weight of $\alpha$ in the objective function defined in Equation~\eqref{eqn:loss} on the attack performance. 
The performance is evaluated using MD (Figure~\ref{fig:p2-client}), JS-divergence (Figure~\ref{fig:p3-client}), and CS (Figure~\ref{fig:p4-client}).
As $\alpha$ increases, the optimization places more emphasis on the $L_1$ term. 
For three datasets (Cora, Citeseer, and Amazon Computer), the performance changes slowly across different $\alpha$ values, which indicates robustness to the loss weighting value. 
However, PubMed exhibits higher sensitivity, where larger $\alpha$ values lead to worse performance in all attack metrics.
This might be because, unlike other datasets, PubMed contains only 3 classes, and a smaller absolute distribution shift might cause larger alignment changes. 
The optimal $\alpha$ value 0.05 is selected based on the observation from all datasets.

\subsection{All Layers vs Last Layer Gradients}

\begin{table*}[tbh]
    \centering 
    \caption{Comparison of attack performance and training time when exploiting all layer gradients and the first layer gradients. 
    The GCN model was used as the target model, and the MLP-based attack model was used to report the results.}
    \begin{tabular}{cl cc cc cc cc}
    \toprule
       \textbf{Dataset}  &
       \textbf{Target label distribution} & 
       \multicolumn{2}{c}{\textbf{Manhattan distance ($\downarrow$)}} & 
         \multicolumn{2}{c}{\textbf{Jensen-Shannon divergence ($\downarrow$)}} & 
         \multicolumn{2}{c}{\textbf{Cosine similarity ($\uparrow$)}} & 
         \multicolumn{2}{c}{\textbf{Time (seconds)}} \\
    \cmidrule(lr){3-4} \cmidrule(lr){5-6} \cmidrule(lr){7-8} \cmidrule{9-10}
    & & AL & LL & AL & LL & AL & LL & AL & LL \\
    \midrule
    \multirow{5}{*}{Cora}
             & Equal proportion & 0.779 & \textbf{0.198} & 0.016 & \textbf{0.001} & 0.662 & \textbf{0.966}  &  \multirow{5}{*}{30.663} & \multirow{5}{*}{\textbf{0.280}}\\
             & Random split & 0.841 & \textbf{0.184} & 0.016 & \textbf{0.002} & 0.627 & \textbf{0.970} \\
             & One class missing & 0.906 & \textbf{0.543} & \textbf{0.064} & 0.130 & 0.726 & \textbf{0.791} \\
             & Single class only & \textbf{0.994} & 1.679 & \textbf{0.265} & 0.483 & \textbf{0.917} & 0.410 \\
             & One-class dominant & 1.490 & \textbf{1.166} & 0.073 & \textbf{0.036} & 0.184 & \textbf{0.641} \\
    \midrule
    \multirow{5}{*}{PubMed}
             & Equal proportion & 0.407 & \textbf{0.333} & 0.011 & \textbf{0.006} & 0.900 & \textbf{0.940} & \multirow{5}{*}{9.699} & \multirow{5}{*}{\textbf{0.213}} \\
             & Random split & 0.524 & \textbf{0.254} & 0.016 & \textbf{0.003} & 0.868 & \textbf{0.964} \\
             & One class missing & \textbf{0.287} & 0.834 & \textbf{0.181} & 0.574  & \textbf{0.974} & 0.676 \\
             & Single class only & 0.928 & \textbf{0.842} & 0.617 & \textbf{0.563} & 0.836 & \textbf{0.844}\\
             & One-class dominant & 0.428 & \textbf{0.383} & 0.013 &\textbf{ 0.019} & 0.929 & \textbf{0.961} \\
    \midrule
    \multirow{5}{*}{Citeseer}
             & Equal proportion & 0.812 & \textbf{0.225} & 0.023 & \textbf{0.001} & 0.665 & \textbf{0.969} & \multirow{5}{*}{17.820} & \multirow{5}{*}{\textbf{0.341}} \\
             & Random split & 0.733 & \textbf{0.133} & 0.018 & \textbf{0.000} & 0.753 & \textbf{0.984} \\
             & One class missing & 0.779 & \textbf{0.552}  & 0.028 &\textbf{ 0.166} & 0.731 & \textbf{0.779} \\
             & Single class only & \textbf{0.886} & 1.547 & \textbf{0.278} & 0.515 & \textbf{0.926} & 0.538\\
             & One-class dominant &1.544 &\textbf{1.306} & 0.103 & \textbf{0.049} & 0.157 & \textbf{0.458}  \\
    \midrule  
    \multirow{5}{*}{AC}
             & Equal proportion & \textbf{0.794} & 0.796 & \textbf{0.011} & 0.012 & 0.674 & \textbf{0.688} & \multirow{5}{*}{183.703} & \multirow{5}{*}{\textbf{0.383}}  \\
             & Random split & 0.278 & \textbf{0.135} & 0.004 & \textbf{0.000} & 0.947 & \textbf{0.993} \\
             & One class missing & 0.298 & \textbf{0.120} & 0.022 & \textbf{0.014} & 0.943 & \textbf{0.993} \\
             & Single class only & 1.755 & \textbf{1.731} & 0.360 & \textbf{0.353} & 0.259 & \textbf{0.291} \\
             & One-class dominant & 0.884 & \textbf{0.800} & 0.012 & \textbf{0.009} & 0.878 & \textbf{0.896} \\     
       \bottomrule
    \end{tabular}
    \label{tab:fl_vs_ll}
\end{table*}

We compare attack performance and training time when training the attack model using (i) all-layer gradients and (ii) last-layer gradients. 
The all-layer gradients are considered because the front layers of GNN primarily capture $k$-hop neighborhood information, which provides structural as well as feature information. 
The last layer was selected as it is connected to the prediction logits and thus carries strong label-specific information through backpropagation, consistent with prior works ~\cite{dai2024decaf, gu2023ldia}. 
In both cases, the attack model is trained on the gradient change, following the same shadow model construction as shown in Equation~\eqref{eqn:grad}. 

As shown in Table~\ref{tab:fl_vs_ll}, using last-layer gradients significantly reduces training time across all datasets, with up to 479× speedup on Amazon Computer, due to the much smaller parameter dimension $W \in \mathbb{R}^{4{,}800}$, compared to the all-layer weight of dimension $W \in \mathbb{R}^{2{,}962{,}400}$. 
Moreover, last-layer gradients achieve better attack performance in nearly all label distribution settings, with only a few exceptions: the \emph{single-class-only} scenario in Cora and Citeseer, the \emph{one-class-missing} case in PubMed, and the \emph{equal-proportion} setting in Amazon Computer. 
Based on both computational efficiency and inference accuracy, we adopt the last-layer gradients for all subsequent experiments.

\subsection{CNN vs MLP-based attack model}
We select a CNN and an MLP-based attack model, as they are lightweight compared to heavy models, such as transformers. 
We compare them in terms of their attack performance, prediction time, and the model's learnable parameters. 
Since the attack data was the last layer weights difference with dimension of $\mathbf{X} \in \mathbb{R}^{Y \times U}$, where Y is the number of classes, and U is the number of neurons in the second last layer, we treat each sample as a long rectangular image. 
The detailed implementation description of CNN is provided in the Supplementary material. 
We observe that with $\sim44$ times fewer parameters than CNN, MLP achieves comparable attack performance (0.985 CS) as shown in Table ~\ref{tab:cnn_vs_nn}.

\begin{table}[t]
    \centering
    \caption{Result comparison between CNN and MLP-based attack model. The numbers reported are for random distribution using GCN as the target GNN model.}
    \scriptsize
    \begin{tabular}{cccc}
    \toprule
    \textbf{Attack model} & \textbf{Cosine similarity} & \textbf{Prediction time (seconds)} & \textbf{Parameters} \\
    \midrule
    MLP & 0.985  & 0.000 & 38,919 \\
    CNN & 0.980 & 0.014 & 1,740,583\\
    \bottomrule 
    \end{tabular}
    \label{tab:cnn_vs_nn}

\end{table}

% \section{Limitation}

\section{Conclusion}
\label{sec:conclusion}
In this work, we investigate the crucial privacy leakage problem in FedGNN settings, termed as the LDI attack.
Specifically, we have introduced Fed-Listing, a novel passive LDI attack against horizontal federated graph learning. 
% By training the shadow FL instances on the honest-but-curious server with the last-layer gradients of clients, Fed-Listing demonstrates strong effectiveness in inferring client label distributions across a variety of benchmark graph datasets and GNN architectures. 
The experimental results show that Fed-Listing consistently outperforms baselines, particularly under challenging non-i.i.d. scenarios such as single-class and one-class-dominant distributions. 
Moreover, we have further evaluated the robustness of Fed-Listing against three widely adopted defense strategies, where the study reveals the difficulty in mitigating Fed-Listing by existing methods, signifying the importance of designing privacy-preserving mechanisms that account for label distribution leakage. 
% One limitation of this work is its reliance on an auxiliary dataset, and we have not evaluated scenarios where such data are unavailable.
% Therefore, exploring the use of synthetic or mixed-source auxiliary data to replace real auxiliary samples is an important future direction. 

\section*{Acknowledgment}
This work was supported by the National Science Foundation under grants No. 2429960 and 2434899, and the Institute of Information \& communications Technology Planning \& Evaluation (IITP) grant funded by the Korea government (MSIT) (No. RS-2024-00431388, the Global Research Support Program in the Digital Field program).

\bibliography{references}
\bibliographystyle{IEEEtran}

\end{document}